\documentclass{article}

\usepackage{microtype}
\usepackage{graphicx}
\usepackage{subfigure}
\usepackage{booktabs}

\usepackage{hyperref}

\usepackage[accepted]{icml2024}

\usepackage{amsmath}
\usepackage{amssymb}
\usepackage{mathtools}
\usepackage{amsthm}

\usepackage[capitalize,noabbrev]{cleveref}

\usepackage[textsize=tiny]{todonotes}

\icmltitlerunning{Accelerating the inference of string generation-based
chemical reaction models for industrial applications}

\begin{document}

\twocolumn[
\icmltitle{Accelerating the inference of string generation-based \\ chemical reaction models for industrial applications}

\begin{icmlauthorlist}
\icmlauthor{Mikhail Andronov}{lug,pfi}
\icmlauthor{Natalia Andronova}{ind}
\icmlauthor{Michael Wand}{lug,hel}
\icmlauthor{Jürgen Schmidhuber}{lug,kau}
\icmlauthor{Djork-Arné Clevert}{pfi}
\end{icmlauthorlist}

\icmlaffiliation{lug}{IDSIA, USI, SUPSI, 6900 Lugano, Switzerland}
\icmlaffiliation{pfi}{Machine Learning Research, Pfizer Research and Development, 10117 Berlin, Germany}
\icmlaffiliation{kau}{AI Initiative, KAUST, 23955 Thuwal, Saudi Arabia}
\icmlaffiliation{hel}{Institute for Digital Technologies for Personalized Healthcare, SUPSI, 6900 Lugano, Switzerland}
\icmlaffiliation{ind}{Independent researcher}

\icmlcorrespondingauthor{Mikhail Andronov}{mikhail.andronov@idsia.ch}

\icmlkeywords{Reaction models, inference speed, transformers}

\vskip 0.3in
]

\printAffiliationsAndNotice{}  
\begin{abstract}
Template-free SMILES-to-SMILES translation models for reaction prediction and single-step retrosynthesis are of interest for industrial applications in computer-aided synthesis planning systems due to their state-of-the-art accuracy. However, they suffer from slow inference speed. We present a method to accelerate inference in autoregressive SMILES generators through speculative decoding by copying query string subsequences into target strings in the right places. We apply our method to the molecular transformer implemented in Pytorch Lightning and achieve over 3X faster inference in reaction prediction and single-step retrosynthesis, with no loss in accuracy.
\end{abstract}

\section{Introduction}
\begin{figure}[ht]
\vskip 0.2in
\begin{center}
\centerline{\includegraphics[width=\columnwidth]{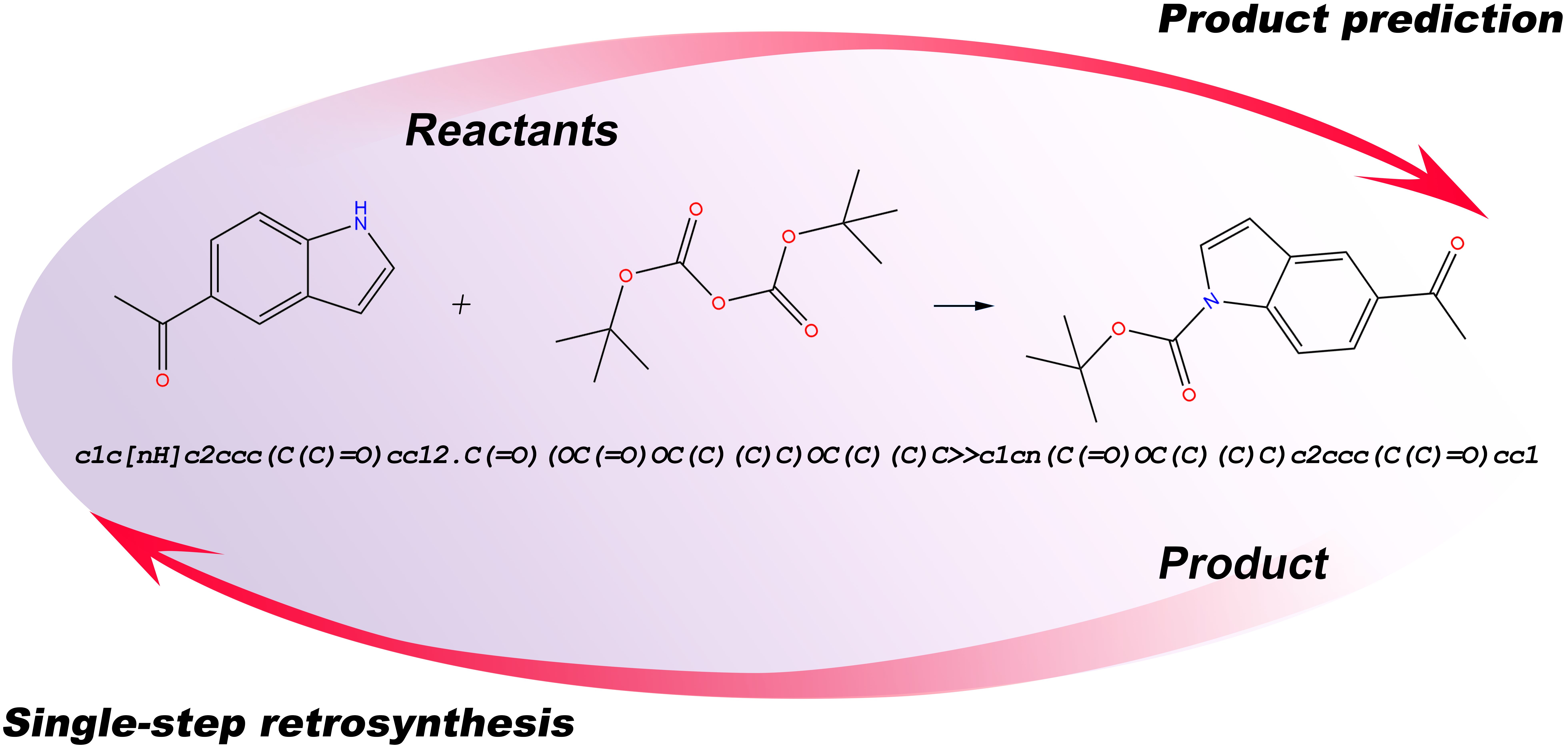}}
\caption{Both reaction product prediction and single-step retrosynthesis can be formulated as SMILES-to-SMILES translation and approached with a model like an encoder-decoder transformer.}
\label{fig1}
\end{center}
\vskip -0.2in
\end{figure}

Automated planning of organic chemical synthesis, first formalized around fifty years ago \cite{Pensak1977}, is one of the core technologies enabling computer-aided drug discovery. While first computer-aided synthesis planning (CASP) systems relied on manually encoded rules \cite{Johnson1989,Gasteiger2000}, researchers now primarily focus on CASP methods powered by artificial intelligence techniques. The design principles of the latter were outlined in the seminal work by Segler et al. \cite{Segler2018}: a machine learning-based single-step retrosynthesis model combined with a planning algorithm. The former proposes several candidate retrosynthetic chemical transformations for a given molecule, and the latter, e.g., Monte-Carlo Tree Search, uses those candidates to construct a synthesis tree.
Single-step retrosynthesis models are now commonly developed in two paradigms: template-based models and template-free models. Besides retrosynthesis, one can also build a model to predict the products of chemical reactions (Figure \ref{fig1}).

The principle of template-based models is to use a set of SMIRKS templates extracted from reaction data and a machine learning model for classification or ranking to select a matching template for a query SMILES that will, upon application, transform the query into the product SMILES (for product prediction), or the SMILES of possible reactants (for single-step retrosynthesis). In contrast, in template-free models, the query transforms into the result without resorting to SMIRKS templates, e.g., with a sequence of predicted graph edits \cite{Sacha2021,Bradshaw2018} or through “translation” of the query SMILES into the desired SMILES with a conditioned text generation model \cite{Schwaller2019,Tetko2020,Irwin2022}. While CASP systems leveraging template-based single-step models proved to be effective \cite{Genheden2020}, there is an interest in building CASP with template-free models instead, as they demonstrate state-of-the-art accuracy in both single-step retrosynthesis and reaction product prediction.
Most accurate template-free models are currently conditional autoregressive SMILES generators based on the transformer architecture \cite{Vaswani2017}, which also serves as the backbone for Large Language Models (LLM) \cite{Brown2020, Zhao2023}. Unfortunately, the high accuracy of autoregressive models like Chemformer \cite{Irwin2022} comes at the cost of a slow inference speed \cite{Torren2024}, which hinders their successful adoption as part of industrial CASP systems.
In our work, we propose a method to accelerate inference from SMILES-to-SMILES translation models based on speculative decoding \cite{Leviathan2023, Xia2023}, a general technique for LLM inference acceleration, combined with insights from the chemical essence of the problem. We reimplement the Molecular Transformer \cite{Schwaller2019} in Pytorch Lightning and use our method to demonstrate its inference acceleration in single-step retrosynthesis and product prediction by up to three times and more without changing the model architecture, training procedure and without compromising on accuracy.

\section{Methods}
\subsection{Algorithm}
Autoregressive models, such as Transformer variants  \cite{Vaswani2017,Brown2020,Schmidhuber1992}, generate sequences token by token, and every prediction of the next token requires a forward pass of the model. Such a process may be computationally expensive, especially for models with billions of parameters. Therefore, an intriguing question arises whether could one generate several tokens in one forward pass of the model, thus completing the output faster. Speculative decoding \cite{Xia2023,Leviathan2023} answers positively. Recently proposed as a method of inference acceleration for Large Language Models, it is based on the draft-and-verify idea. At every generation step, one can append some draft sequence to the sequence generated by the model so far and see if the model "accepts" the draft tokens.

\begin{figure*}
\vskip 0.2in
\begin{center}
\centerline{\includegraphics[width=1.0\textwidth]{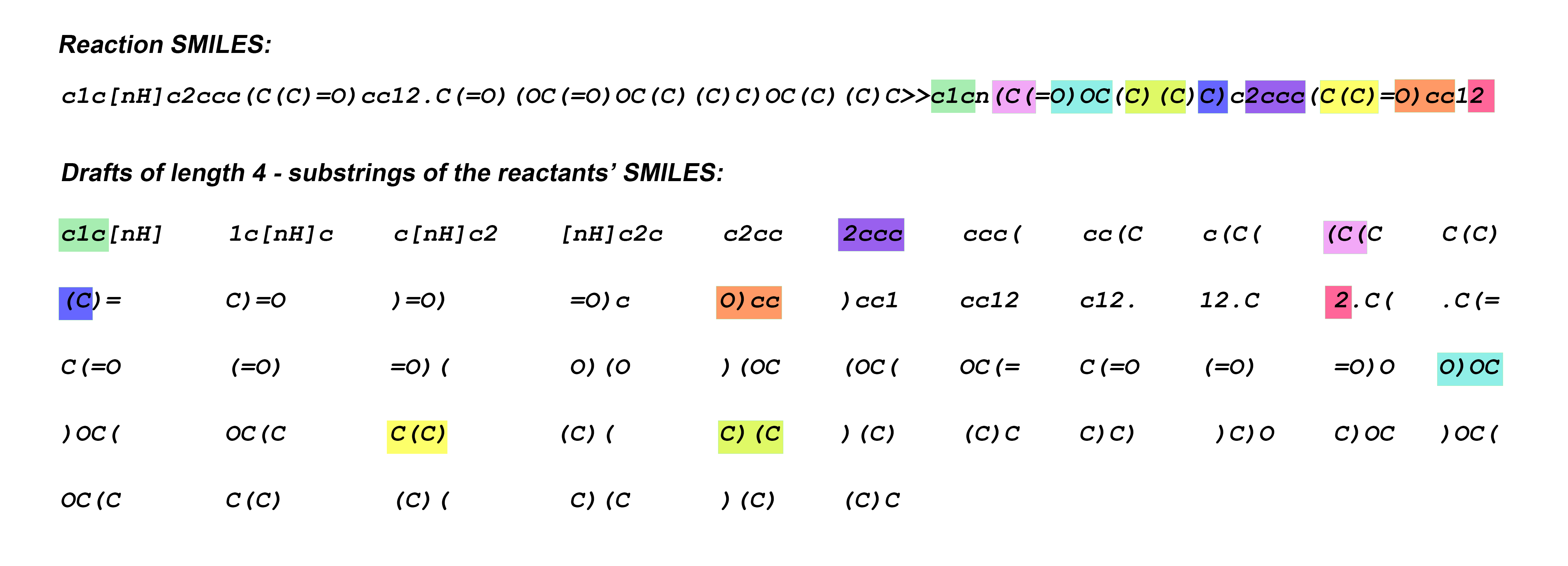}}
\caption{Speculative decoding accelerates product prediction with the molecular transformer or a similar autoregressive SMILES generator. Before generating an output sequence, we prepare a list of subsequences of a desired length, e.g., four, of the tokenized query SMILES of reactants. Then, at every generation step, the model can copy up to four tokens from one of the draft sequences to the output, thus generating from one to five tokens in one forward pass.}
\label{fig2}
\end{center}
\vskip -0.2in
\end{figure*}

If the draft sequence has length $N$, in the best case, the model adds $N + 1$ token to the generated sequence in one forward pass, and in the worst case, it adds one token as in standard autoregressive generation. The acceptance rate for one generated sequence is the number of accepted draft tokens divided by the total number of tokens in the generated sequence. One can also test multiple draft sequences in one forward pass taking advantage of parallelization, and choose the best one. Speculative decoding does not affect the content of the predicted sequence compared to the token-by-token decoding in any way.  

One can freely choose a way of generating draft sequences. For LLMs, one would usually use another smaller language model that performs its forward pass faster than the main LLM \cite{Leviathan2023} or additional generation heads on top of the LLM's backbone \cite{Cai2024}. However, one can also construct draft sequences without calling any learned functions. For example, generate random draft sequences, even though their acceptance rate will be minimal, or assemble draft sequences out of tokens in the query sequence — a prompt for decoder-only language models or a source sequence for translation models. The latter option is perfect for retrosynthesis or reaction prediction as SMILES-to-SMILES translation. In a chemical reaction, large fragments of reactants typically remain unchanged, which means that the SMILES of products and reactants have many common substrings. It is especially true if reactant and product SMILES are aligned to minimize the edit distance between them \cite{Zhong2022}. Therefore, we can extract multiple substrings of a chosen length $N$ from the query SMILES and use them as draft sequences with a high acceptance rate. Figure \ref{fig2} demonstrates this method in product prediction. Before generating the target string, we assemble a list of token subsequences from the source sequence (reactant tokens) with a sliding window of a desired length (4 in this case) and stride 1. Then, at every generation step, we can try all draft sequences in one forward pass of the model to see if the model can copy up to 4 tokens from one of them. The draft token acceptance rate in this example reaches 78\%. 

Speculative decoding does not require any changes to the model architecture or training of additional models. The cost of generating draft sequences in this way is negligible compared to that of the forward pass of the reaction model.

\subsection{Model}
We demonstrate the application of our method to the Molecular Transformer \cite{Schwaller2019}. It is an encoder-decoder transformer model suitable for SMILES-to-SMILES translation. We conduct our experiments on one H100 GPU with 80 GB memory.

The original Molecular Transformer \cite{Schwaller2019} adopts OpenNMT \cite{Klein2018}, a general framework for neural machine translation, for SMILES-to-SMILES translation. Since the code in this framework is complex and intractable to customize, we decided to re-implement the model in PyTorch Lightning to keep only the necessary code and have more design freedom in the model's inference procedure implementation.

\begin{table}[t]
\caption{The top-5 accuracy of the Molecular Transformer (MT) in predicting reaction products on USPTO MIT (mixed) in the original report and our reimplementation. Both models use beam search with beam size 5 to generate five predictions for every query. Our model successfully reproduces the accuracy of the original model with negligible discrepancy.}
\label{table_forw_acc}
\vskip 0.15in
\begin{center}
\begin{small}
\begin{sc}
\begin{tabular}{lccc}
\toprule
Accuracy & Original MT & Our MT & $\Delta$ \\
\midrule
Top-1, \%   & 88.6 & 88.4 & -0.2 \\
Top-2, \%   & 92.4 & 92.5 & 0.1\\
Top-3, \%   & 93.5 & 93.7 & 0.2 \\
Top-5, \%   & 94.5 & 94.7 & 0.2 \\
\bottomrule
\end{tabular}
\end{sc}
\end{small}
\end{center}
\vskip -0.1in
\end{table}

\subsection{Data}
We used the open reaction data from US patents \cite{Lowe2012} for training all models. We trained the model for reaction product prediction as in the original paper \cite{Schwaller2019} on the USPTO MIT dataset, a standard benchmark for product prediction, without reactant-reagent separation. We trained the model for single-step retrosynthesis on USPTO 50K, a standard dataset for the task. In this dataset, we augmented every reaction in the training set 20 times using SMILES augmentation \cite{Tetko2020} with root-aligned SMILES \cite{Zhong2022}. We followed the standard atomwise tokenization procedure \cite{Schwaller2019} to tokenize SMILES.

\section{Results and Discussion}
Our implementation of the Molecular Transformer (MT) successfully reproduces the accuracy scores of the original MT \cite{Schwaller2019} that relies on OpenNMT. Comparing our MT and the original MT, we observe at most 0.2 percentage points discrepancy of top-1 to top-5 accuracy in product prediction with beam search (Table \ref{table_forw_acc}). Having verified the correctness of our MT's implementation in this way, we replace the standard decoding procedures for the MT with our methods based on speculative decoding and achieve a significant speed-up in product prediction and single-step retrosynthesis.

\subsection{Product prediction}
We tested our MT for product prediction on USPTO MIT mixed, i.e., without an explicit separation between reactants and reagents. The test dataset in this benchmark comprises 40 thousand reactions.

When serving a reaction prediction model as an AI assistant for chemists,  one could use greedy decoding with a batch size of 1 for inference. Table \ref{table_forw_greedy} summarizes our experiments with greedy decoding from MT on the test set of USPTO MIT. The model's inference with standard greedy decoding with a batch size of 1 finishes in around 62 minutes. In contrast, if we use greedy generation enhanced with our speculative decoding, the inference time reduces to 26 minutes with a draft length of 4 and 17 minutes with a draft length of 10, which corresponds to 137 \% and 262 \% speedup, respectively. The acceptance rate in our drafting method averaged over all test reactions is 79\%. Potentially, it can be even higher if one adds more draft sequences to choose from, for example, subsequences of the source sequence dilated by one token. Of course, greedy decoding with a large batch size is much faster and completes in around 4 minutes with 32 reactions in a batch. However, accelerating inference with a batch size of 1 would be sufficient for an improved user experience with reaction prediction assistants: chemists would usually enter one query at a time in a user interface of a reaction model like IBM RXN. The model's accuracy is 88.3\% with both standard and speculative greedy decoding.

\begin{table}[t]
\caption{Wall time of the model's inference on the USPTO MIT test set in reaction product prediction with standard and speculative greedy decoding. B stands for batch size, and DL stands for draft length. The time is averaged over five attempts.}
\label{table_forw_greedy}
\vskip 0.15in
\begin{center}
\begin{small}
\begin{sc}
\begin{tabular}{lcc}
\toprule
Decoding & Time, min \\
\midrule
Greedy (B=1)   & 61.8 $\pm$ 5.88 \\
Greedy speculative (B=1, DL=4)   & 26.04 $\pm$ 2.07 \\
Greedy speculative (B=1, DL=10)   & 17.06 $\pm$ 0.25 \\
Greedy (B=32)   & 4.13 $\pm$ 0.06 \\
\bottomrule
\end{tabular}
\end{sc}
\end{small}
\end{center}
\vskip -0.1in
\end{table}

\begin{table}[t]
\caption{Wall time of the model’s inference on the USPTO 50k
test set (5K reactions) in single-step retrosynthesis with beam search (BS) and speculative
beam search (SBS). Batch size is 1. DL stands for draft length, and beam width is denoted as $n$. The time is averaged over five attempts.}
\label{table_retro_specul_time}
\vskip 0.15in
\begin{center}
\begin{small}
\begin{sc}
\begin{tabular}{lccc}
\toprule
Decoding & $n$=5, min & $n$=10, min & $n$=25, min \\
\midrule
BS   &  36.7 $\pm$ 0.3 & 39.9 $\pm$ 0.3 & 46.2 $\pm$ 0.3 \\
SBS, DL=10  & 9.9 $\pm$ 0.1 & 15.4 $\pm$ 0.1 & 28.1 $\pm$ 0.1 \\
SBS, DL=0  & 23.1 $\pm$ 0.3 & 25.7 $\pm$ 0.3 & 34.6 $\pm$ 0.3 \\
\bottomrule
\end{tabular}
\end{sc}
\end{small}
\end{center}
\vskip -0.1in
\end{table}

\subsection{Single-step retrosynthesis}
We carried out single-step retrosynthesis experiments on USPTO 50K, in which the training dataset was augmented 20 times. The augmentation procedure is to construct alternative root-aligned \cite{Zhong2022} SMILES for every dataset entry. This augmentation minimizes the edit distance between reactants and products, which simplifies training, pushes the model's accuracy higher, and increases the acceptance rate in our speculative decoding method. Speculative decoding in single-step retrosynthesis accelerates greedy decoding as much as in reaction product prediction. However, this has limited utility. In synthesis planning, one would want a single-step retrosynthesis model to suggest multiple different reactant sets for every query product so that the planning algorithm can choose from them. Usually, one would employ beam search to generate several outputs from the transformer for single-step retrosynthesis. 

We found a way to accelerate beam search with speculative decoding. The main idea behind it is that at every decoding step, we use a draft sequence to generate multiple candidate sequences of different lengths in one forward pass, which we then order by probabilities and keep the best $n$ of them. To organize such sequences of unequal lengths in a batch for a subsequent forward pass of the model, we pad them from the left and offset the positional encodings accordingly for every sequence. We call our algorithm "speculative beam search" (SBS). The full description of the algorithm is in the Appendix \ref{appendixB}.

\begin{table}[t]
\caption{The top-N accuracy of our single-step retrosynthesis model on USPTO 50K with beam search (BS) and speculative beam search (SBS). DL stands for draft length.}
\label{table_acc_retr}
\vskip 0.15in
\begin{center}
\begin{small}
\begin{sc}
\begin{tabular}{lccc}
\toprule
Accuracy & BS & SBS, DL=10 & SBS, DL=0 \\
\midrule
Top-1, \%   & 52.07 & 52.07 & 52.07 \\
Top-3, \%   & 75.16 & 75.16 & 75.16 \\
Top-5, \%   & 82.07 & 82.07 & 82.07 \\
Top-10, \%   & 89.08 & 89.08 & 89.08 \\
Top-25, \%   & 94.07 & 94.05 & 94.09 \\
\bottomrule
\end{tabular}
\end{sc}
\end{small}
\end{center}
\vskip -0.1in
\end{table}

The wall time our retrosynthesis model takes to process the USPTO 50K test set with beam search and batch size 1 is around 37 minutes, 40 minutes, and 46 minutes when generating 5, 10, and 25 predictions for every query SMILES, respectively (Table \ref{table_retro_specul_time}). We keep the beam width and the number of best sequences equal. When we replace the standard beam search with our SBS with the draft length of ten, the generation finishes in around 10, 15, and 28 minutes, accelerating the inference by around 3.7, 2.7, and 1.8 times, respectively. 

SBS reduces to the standard beam search when draft tokens are never accepted. It happens, for example, if we use the start-of-sequence token as a single draft. The "effective" batch size also does not increase when using only one draft per forward pass (see Section \ref{limitations}). We added an experiment in this mode, denoting it as "SBS, DL=0" (Tables \ref{table_retro_specul_time} and \ref{table_acc_retr}), to verify that the algorithm's acceleration comes from the draft sequences. Interestingly, SBS is faster than the standard beam search even in case of DL=0 (Table \ref{table_retro_specul_time}).

The top-N accuracy when using SBS is practically the same as the accuracy of the standard beam search (Table \ref{table_acc_retr}). Only the difference in the top-25 accuracy is visible, although tiny, only a couple hundredths of a percent point.
Thus, our SBS accelerates the MT's generation of multiple reactant sets several times without having to compromise on accuracy at all. Such a speed-up could make the transformer a more attractive single-step model for multi-step synthesis planning.

\subsection{Limitations}\label{limitations}
The speed of the model's forward pass decreases with the increase of the size of a batch input to the transformer decoder. This effect quickly manifests itself in our drafting strategy. The latter is "brute force" in some sense, as we use various substrings of the source SMILES as drafts in parallel at every forward pass of the model. It inflates the "effective" batch size of the input to the transformer decoder: we copy every sequence in the batch as many times as there are drafts and concatenate the sequence copy and the corresponding draft. As a result, the generation of target SMILES may become slower with speculative decoding than with standard decoding procedures, even when it requires fewer calls to the model. We put a boundary on the number of drafts extracted from a query sequence to mitigate this deceleration. However, this compromises the acceptance rate.

The speed of the model's forward pass decreases with the increase of the size of a batch input to the transformer decoder. This effect quickly manifests itself in our drafting strategy. The latter is "brute force" in some sense, as we use various substrings of the source SMILES as drafts in parallel at every forward pass of the model. It inflates the "effective" batch size of the input to the transformer decoder: we copy every sequence in the batch as many times as there are drafts and concatenate the sequence copy and the corresponding draft. As a result, the generation of target SMILES may become slower with speculative decoding than with standard decoding procedures, even when it requires fewer calls to the model. We put a boundary on the number of drafts extracted from a query sequence to mitigate this deceleration. However, this compromises the acceptance rate.
In addition, as the size of the input batch grows, the generation speed with speculative decoding becomes bottlenecked in terms of the number of calls to the model. The "least lucky" sequence with the lowest acceptance rate of speculative tokens would determine the number of calls.
Together, these two effects limit the utility of our speculative decoding for large batch sizes and beam widths. For example, our SBS is slower than the standard beam search when the beam size is fifty. Nonetheless, we consider speculative decoding very promising for synthesis planning acceleration, as it significantly improves the speed of the standard beam search without losing accuracy in the practical range of beam widths ($\sim$10-20).

Designing a drafting strategy for SMILES that removes the need for multiple parallel drafts while retaining a high acceptance rate is an aspect of our ongoing work.

Our speculative algorithm works well in single-step retrosynthesis because the MT typically predicts low-entropy next-token distributions in this task. In most cases, all predicted probability mass concentrates on one token. Consequently, long sequences often win in the probability competition against shorter sequences at every speculative beam search iteration. If several top probabilities from the predicted next-token distribution are approximately the same, then short sequences would tend to have higher probabilities, and the model would generate only a small number of tokens per forward pass, benefitting little from drafts.

\section{Conclusion}
We combine speculative decoding and chemical insights to accelerate inference in the molecular transformer, a SMILES-to-SMILES translation model. Our method makes processing the test set more than three times faster in both single-step retrosynthesis on USPTO 50K and reaction product prediction on USPTO MIT compared to the standard decoding procedures. Our method aims at making state-of-the-art template-free SMILES-generation-based models such as the molecular transformer more suitable for industrial applications such as computer-aided synthesis planning systems.

\section*{Acknowledgements}
This study was partially funded by the European Union’s Horizon 2020 research and innovation program under the Marie Skłodowska-Curie Innovative Training Network European Industrial Doctorate grant agreement No. 956832 “Advanced machine learning for Innovative Drug Discovery, and also by the Horizon Europe funding programme under the Marie Skłodowska-Curie Actions Doctoral Networks grant agreement “Explainable AI for Molecules - AiChemist” No. 101120466. 

\section*{Code availability}
The code and checkpoints for trained models will be available at \url{https://github.com/Academich/translation-transformer}.

\section*{Conflict of interests}
The authors have no
conflicts of interests.

\bibliography{paper}
\bibliographystyle{icml2024}

\newpage
\appendix
\onecolumn
\section{Training details.}
For product prediction, we train this model with the same hyperparameters as in \citeauthor{Schwaller2019} with four encoder and decoder layers, eight heads, embedding dimensionality of 256, and feedforward dimensionality of 2048, which results in 11,4 million parameters. For single-step retrosynthesis, we set the hyperparameters as in \citeauthor{Zhong2022} (six encoder and decoder layers, eight heads, embedding dimensionality of 256, and feedforward dimensionality of 2048), which results in 17,4 million parameters. The dictionary is the same for the encoder and the decoder in both models. We use the Adam optimizer for both models.

\section{Speculative Beam Search}\label{appendixB}
The outline of the speculative beam search procedure we implement for our experiments is as follows:

\begin{algorithm}
   \caption{Speculative beam search}
   \label{alg:sbs}
\begin{algorithmic}
   \STATE {\bfseries Input:} Query token sequence (tensor of integers) $s$, number of best sequences $n$
   \STATE $memory = encoder(s)$
   \STATE $drafts = getDrafts(s)$
   \STATE $res = [BOS]$
   \REPEAT
   \STATE $draftedSequences = concatDraftsToSequences(res, drafts)$ 
   \STATE $logits = decoder(draftedSequences, memory)$
   \STATE $bestDraftedSequences = selectBestDraft(logits)$
   \STATE $candidates = sample(bestDraftedSequence)$
   \STATE $bestCandidates = sortAndExtract(candidates, n)$
   \STATE $res = padLeft(bestCandidates)$
   \UNTIL{maximal length or number of iterations is reached}
\end{algorithmic}
\end{algorithm}

\section*{Functions}
\textbf{getDrafts:} Produces a list of subsequences of the query sequence using a sliding window of a given length with stride one. Only the first $N_d$ drafts are kept and the others are discarded to mitigate increasing memory requirements as the number of drafts grows. $N_d$ is typically around 25. 

\textbf{encoder:} Forward pass of the encoder of the Molecular Transformer.

\textbf{decoder:} Forward pass of the decoder of the Molecular Transformer.

\textbf{concatDraftsToSequences}: Concatenates every draft to every resulting sequence decoded so far. This gives a tensor of sequences with the batch size of length($res$) $\times$ length ($drafts$). 
\begin{algorithmic}
    \STATE \textbf{Input:} decoded sequences $res$, tensor of drafts $drafts$
       \STATE $draftedSequences = []$ 
   \FOR{$r$ {\bfseries in} $res$}
   \FOR{$d$ {\bfseries in} $drafts$}
   \STATE $draftedSequences.append( r \mathbin\Vert d )$
   \ENDFOR
   \ENDFOR
   \STATE \textbf{return} $draftedSequences$
\end{algorithmic}

\textbf{selectBestDraft:}
Uses the logits predicted by the decoder to select the best draft for every resulting sequence and discard the others. The best draft is the one with the largest number of accepted tokens.

\begin{figure*}
\vskip 0.2in
\begin{center}
\centerline{\includegraphics[width=1.0\textwidth]{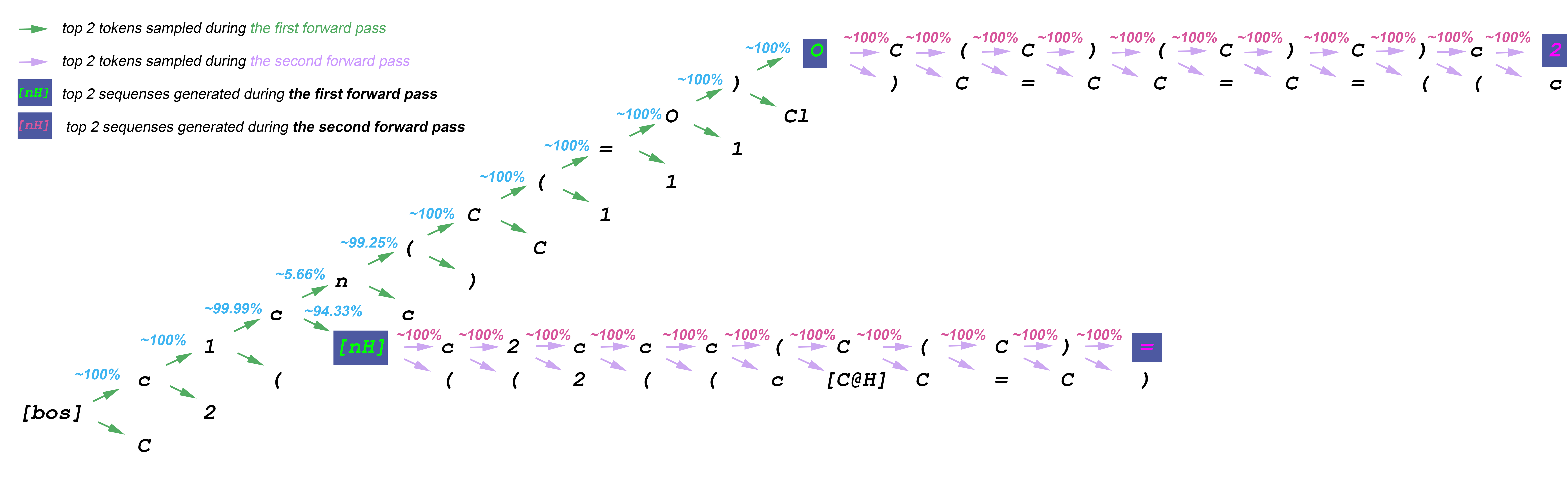}}
\caption{An example of two first iterations of the sampling of candidate sequences in speculative beam search. Here, we select the two best candidates at each iteration. The first forward pass generates 12 candidate sequences. The second forward pass generates 24 sequences. The draft length in this example is 10. The best sequences in the first iteration are \textbf{c1c[nH]} and \textbf{c1cn(C(=O)O}. The best sequences after the second iteration are \textbf{c1c[nH]c2ccc(C(C)=} and \textbf{c1cn(C(=O)OC(C)(C)C)c2}.}
\label{fig:sbs_sample}
\end{center}
\vskip -0.2in
\end{figure*}

\textbf{sample:} Proposes many candidate sequences for the decoding step based on the single accepted draft. The candidate sequences may have different lengths. Figure \ref{fig:sbs_sample} demonstrates an example.

\textbf{sortAndExtract:} Orders the candidate sequences by probabilities, keeps the $n$ ones with the highest probabilities and discards the others.

\textbf{padLeft}: If the best candidate sequences have different lengths, organizes them in a tensor by padding them with a special PAD token from the left. In the next step of the algorithm, the starting positions for the positional encodings in the transformer decoder get shifted accordingly.

\end{document}